\documentclass[twocolumn,10pt]{asme2ej}

\usepackage{amsmath}
\DeclareMathOperator*{\argmin}{arg\,min}
\usepackage{graphicx} 
\usepackage{hyperref}   
\hypersetup{
	colorlinks=true,
	linkcolor=blue,
	citecolor=blue,
	urlcolor=blue,
}
\usepackage[square,numbers]{natbib}

\title{LinFlo-Net: A two-stage deep learning method to generate simulation ready meshes of the heart}

\author{Arjun Narayanan \thanks{Address all correspondence to this author.} \\
    \affiliation{
	Department of Mechanical Engineering\\
	University of California, Berkeley\\
	Berkeley, California 94704\\
    Email: arjun.narayanan@berkeley.edu
    }	
}

\author{Fanwei Kong \\
    \affiliation{
        Department of Pediatrics\\
	Stanford University\\
	Stanford, California 94305\\
	Email: fwkong@stanford.edu
    }
}

\author{Shawn Shadden \\
    \affiliation{
        Department of Mechanical Engineering\\
	University of California, Berkeley\\
	Berkeley, California, 94704\\
	Email: shadden@berkeley.edu
    }
}

\begin{document}

\maketitle    

\begin{abstract}
{\it We present a deep learning model to automatically generate computer models of the human heart from patient imaging data with an emphasis on its capability to generate thin-walled cardiac structures. Our method works by deforming a template mesh to fit the cardiac structures to the given image. Compared with prior deep learning methods that adopted this approach, our framework is  designed to minimize mesh self-penetration, which typically arises when deforming surface meshes separated by small distances. We achieve this by using a two-stage diffeomorphic deformation process along with a novel loss function derived from the kinematics of motion that penalizes surface contact and interpenetration. Our model demonstrates comparable accuracy with state-of-the-art methods while additionally producing meshes free of self-intersections. The resultant meshes are readily usable in physics based simulation, minimizing the need for post-processing and cleanup.
}
\end{abstract}



\section{Introduction}

Image-based computer modeling is playing an increasing role in understanding the mechanisms of cardiac disease and personalized care \cite{prakosa2018personalized}. Broadly, this paradigm uses medical imaging, such as computed tomography (CT) or magnetic resonance (MR), to construct an anatomically accurate computer model of the heart in order to mathematically model physiological processes and probe functional information \cite{corral2020digital}. Reconstructing an accurate, personalized computer model of the heart is challenging because of imaging artifacts, limited resolution and difficultly differentiating between cardiac and surrounding tissues. Further, generating these models manually can take on the order of 6-10 hours \cite{zhuang2019evaluation} for an expert human. This is one of the major hurdles in the adoption of such technologies on a larger scale, motivating the development of automatic and scalable methods.

Segmentation is the process of identifying structures of interest in an image. Recent advances in machine learning and computer vision have demonstrated considerable success in the field of medical image segmentation. Numerous methods have been proposed that can achieve human-level performance on a large variety of structures of interest for the medical community \cite{wasserthal2022totalsegmentator}. However, segmentation often generates artifacts that are unfit for simulation-based modeling. Recently, we have developed alternative template based deep-learning methods that are able to generate simulation-ready computer models of cardiac structures automatically from images \cite{kong2021deep,Kong2022LearningWH}. These methods use machine learning to deform mesh templates to create a personalized geometry that best matches the image data. However, these methods do not guarantee a bijective mapping between the template and the deformed meshes. Thus, self-intersections and unphysiological distortions are possible, requiring significant post-processing steps to correct these artefacts.

\begin{figure}[t]
    \centering
    \includegraphics[width=0.5\textwidth]{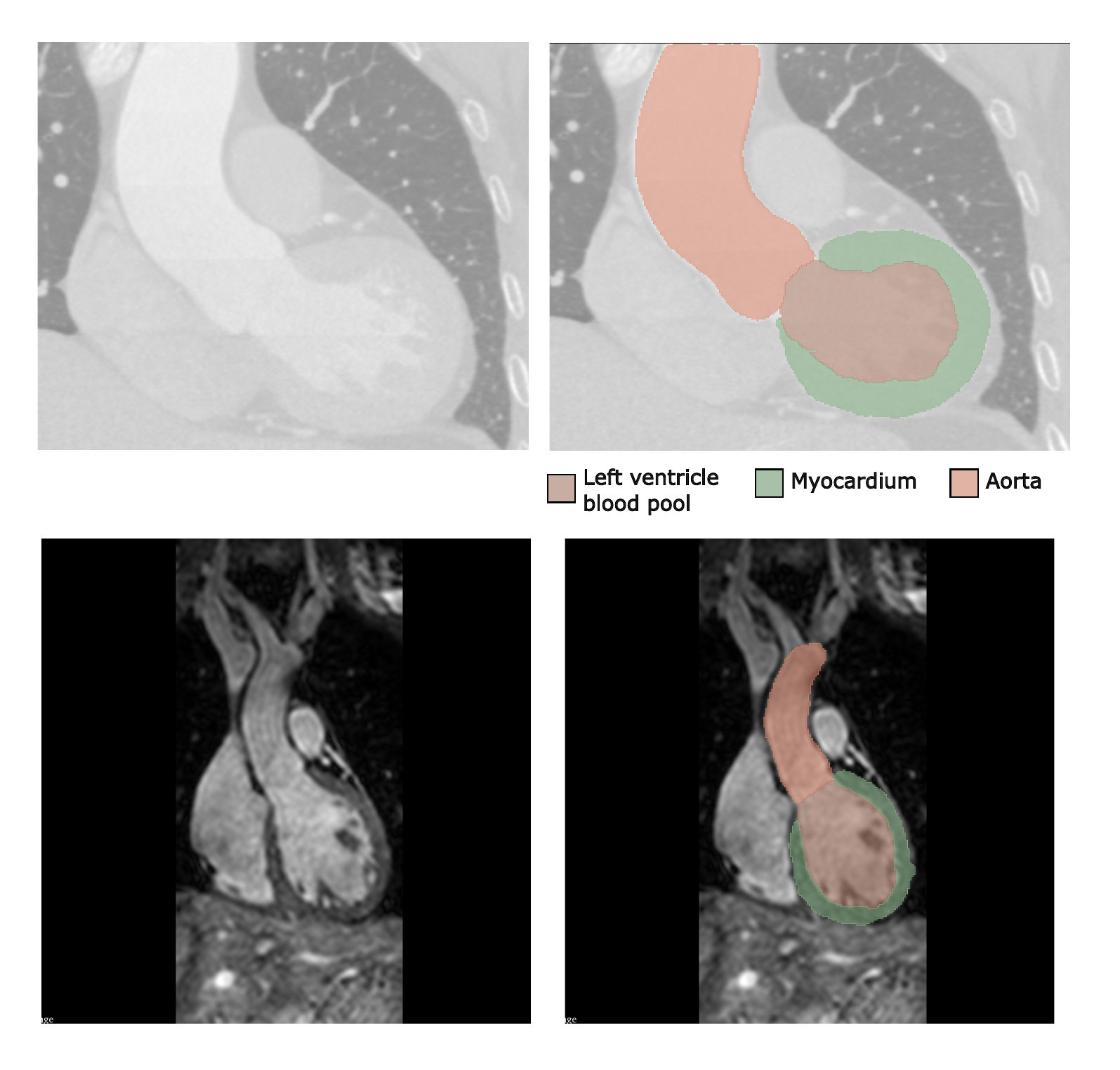}
    \caption{Illustrative examples of CT (top row) and MRI (bottom row) images of the cardiac region. The segmentation of some cardiac structures of interest are overlaid on the figures on the right. Since the myocardium is a thick muscular structure, it is clearly visible in the image. However, the tissue thickness of other structures like the aorta are not visible in these images, and only the blood pool within these structures is visible.}
    \label{fig:tissue-thickness}
\end{figure}

Notably, prior methods have focused primarily on generating surface models of the blood pool boundaries since, except for the left ventricular (LV) myocardium, only blood pools are discerned from clinical imaging (cf.\ Fig.~\ref{fig:tissue-thickness}). However, many applications of cardiac modeling require modeling of the cardiac and vascular tissue. These tissues (with the exception of the LV wall) have modest thickness, and thus deformation of templates that contain such tissue structures are highly susceptible to self-intersections. 

To address those limitations, we present here a deep learning method to produce whole-heart meshes with thin-walled structures from medical image (CT or MR) data. We employ a template-based method to ensure accurate and simulation compatible models free of self-intersections. Briefly, we use deep learning to deform a template mesh using a combination of linear-transformations and diffeomorphic flow deformations. We present a novel physics-based loss term that penalizes vanishing volumes thereby preventing self-penetrations when mapping thin walled structures. The model can be trained on data containing no thickness information and subsequently evaluated on a template with thickness. We demonstrate that this approach is able to successfully deform template meshes in a realistic manner without interpenetration. The predictions of our model can be readily used to generate volumetric grids for computational simulations.

\section{Related Work} \label{sec:related-work}

Automatically generating 3D meshes from images is a challenging problem that has attracted considerable interest recently. In Pixel2Mesh~\cite{wang2020pixel2mesh}, the authors deform an initial ellipsoidal template mesh to a target shape. Voxel2Mesh \cite{wickramasinghe2020voxel2mesh} extended these ideas to 3D medical images wherein the resultant mesh was dynamically refined in areas to capture geometric details. MeshDeformNet~\cite{kong2021deep} and HeartDeformNet \cite{Kong2022LearningWH} demonstrated state-of-the-art accuracy in whole-heart mesh generation while preserving anatomical accuracy. However, these methods are not specifically designed to avoid self-intersections, which hinders their use for physics-based simulations. For example, standard meshing softwares like TetGen~\cite{hang2015tetgen} fail to produce volumetric tetrahedral meshes from surface meshes that contain self-intersections. 

To overcome these challenges, there has been interest in incorporating diffeomorphic constraints to deep learning algorithms, either explicitly through regularization losses or implicitly through inherent network designs, to produce deformation fields that are diffeomorphic. Pak et al.~\cite{pak2021distortion} generated 3D volumetric meshes of the aortic valve by training a neural network to predict the displacement field of an initial template. They proposed a novel distortion energy based on the singular value decomposition of the deformation gradient that penalizes deformations that are not diffeomorphic. The NeuralMeshFlow framework~\cite{gupta2020neural} used a neural network to predict the 3-dimensional vector field that could be used to integrate a point cloud (e.g. the vertices of a mesh) to a target geometry. Under regularity assumptions, this deformation is diffeomorphic. Similar ideas have been extended to applications in 3D medical imaging. CorticalFlow~\cite{lebrat2021corticalflow} and CortexODE~\cite{ma2022cortexode} leveraged diffeomorphic flow fields for cortical surface reconstruction. UNetFlow~\cite{bongratz2023meshes} used a similar approach to generate meshes of abdominal structures such as the liver and pancreas. In practice, the meshes generated by such methods are not strictly intersection free due to various factors like numerical integration errors, but nevertheless, these methods typically show a significant reduction in the number of self-intersecting faces.

The method presented herein was inspired by ideas from the approaches described above. However, our method is unique in several key ways highlighted below:

\begin{enumerate}
    \item We employ a two-stage deformation process consisting of an initial linear transformation followed by diffeomorphic flow-based deformation to capture finer details.
    \item Along with the clinical image, we provide a representation of the template mesh's current position as input to our model. We use an unsigned distance function as the representation since computer vision models are well suited to this format. We observed empirically that this significantly improves model performance.
    \item We demonstrate a simple and effective strategy to minimize integration errors by constraining the magnitude of the flow field, drawing connections to the Courant--Friedrichs--Lewy condition in numerical analysis.
    \item We propose a physics-based loss function derived from the kinematics of motion of continua that penalizes flow fields that generate collapsing volumes, thus preserving structure thickness even when thickness is not visible in the original image.
\end{enumerate}

\section{Methods} \label{sec:methods}

\subsection{Neural Network Architecture}

\begin{figure}[t]
    \centering
    \includegraphics[width=0.47\textwidth]{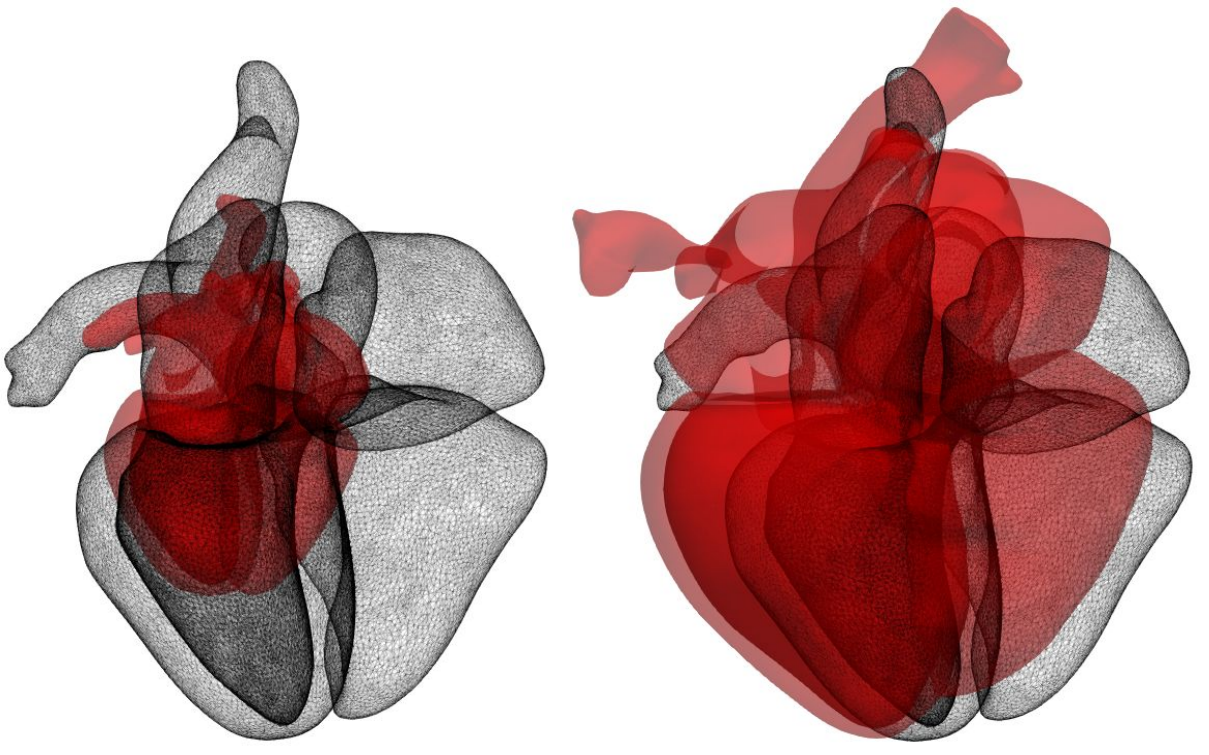}
    \caption{Two samples from our dataset that demonstrate the variation in scale across samples. Black wireframe is the template mesh, and red surface is the ground-truth mesh.}
    \label{fig:scale-difference}
\end{figure}

We aim to create a machine learning method that can deform a template heart model to match with a patient's image data. There can be significant variability in heart geometry. Figure \ref{fig:scale-difference} shows the variation in scale for two hearts in the dataset we will consider. The differences in scale are due to different field of view between scans and inter-patient variation in heart-size. We expect such variations to exist in real-world applications and our model is designed to handle these cases. Namely, our method first performs a linear transformation (cf. sec. \ref{sec:linear-transform}) via scaling, rotation, and translation. A linear transformation is well behaved and guaranteed to be diffeomorphic. Further, it is highly interpretable. Thus, we aim to train our linear deformation module to maximally capture large-scale deformations. A subsequent (nonlinear) mesh deformation module is used to deform the linearly transformed mesh by integrating the mesh vertices along a learned vector field. This approach was designed with the requirement that the deformed template has minimal self-intersections. Since integration errors accumulate over larger deformations, by utilizing the linear transformation module in the first step, we reduce errors and self-intersections. Namely, the flow module is only required to capture finer details. Indeed, we have verified that the quality of meshes produced by the flow module alone (without any linear transformation) is poor.

\subsubsection{Linear transformation.} \label{sec:linear-transform}

A linear transformation is a global operation applied to all mesh vertices. The transformation can be defined by 9 parameters: 3 scaling, 3 rotation, and 3 translation. We observed empirically that scaling and translation have a bigger impact on accuracy. Rotation only provides a small benefit, but we include it nonetheless since there is no significant overhead to do so. Our linear transformation module consists of a 3D convolutional neural network (CNN) encoder and a multi-layer perceptron (MLP) decoder. The CNN consists of multiple layers of convolution followed by downsampling. The CNN processes a normalized input image of size $128^3$ (see sec. \ref{sec:dataset} for details on the dataset and normalization protocol) and produces an encoding of dimension $4^3$ with $512$ channels. This encoding is flattened and processed by an MLP with a single hidden layer to produce the 9 parameters defining a linear transformation. The model is initialized to produce the identity transformation.

\begin{figure}[t]
    \centering
    \includegraphics[width=0.47\textwidth]{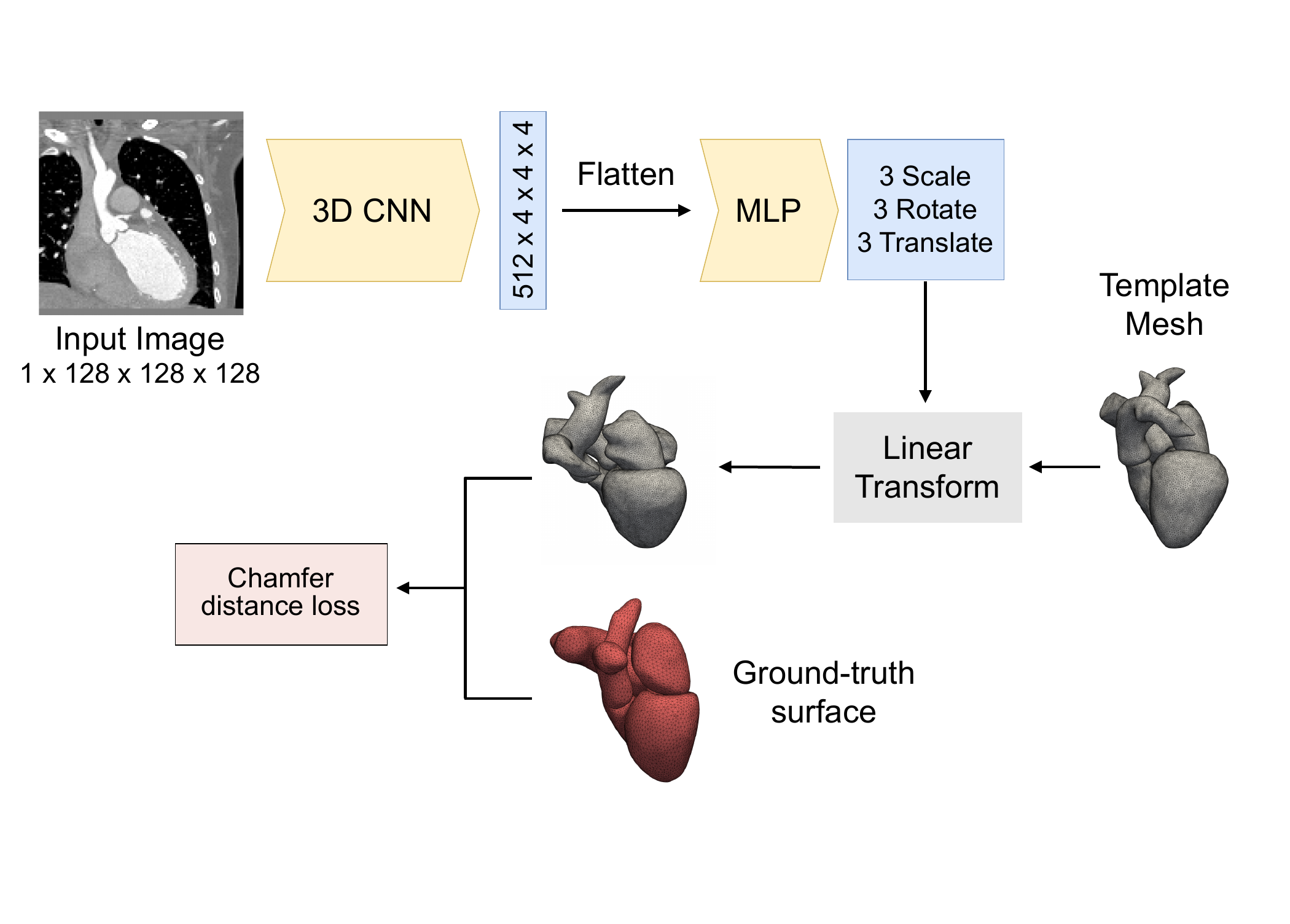}
    \caption{Workflow describing the training pipeline for the linear transformation module}
    \label{fig:linear-transform}
\end{figure}

The predicted parameters are used to perform a linear transformation on a template mesh. We use the center of the image as the origin of the transformation. We apply the operation in the following order: scale--rotate--translate. We use the same initial template, and always initialize it at the same location in normalized image coordinates. The model is trained to minimize the chamfer distance in the L1-norm between the transformed template and ground truth mesh vertices given by eq. \ref{eq:chamfer}. We compute the Chamfer loss for each cardiac structure separately and take its average. The linear transformation and loss function are implemented in PyTorch3D \cite{ravi2020pytorch3d}. We show a schematic of this module in fig. \ref{fig:linear-transform}.

\subsubsection{Flow deformation.} \label{sec:flow-deformation}

The linear transformation is a global operation and is unable to adjust the template to the finer features of the target geometry. We instead require a spatially varying deformation field. While this deformation field can be directly learned, it will generally not be diffeomorphic. Instead, we can learn a flow vector field that can be used to integrate the vertices of the linearly transformed template, resulting in a diffeomophic deformation. The learned vector field is further constrained using the loss function eq. \ref{eq:volume-loss} to prevent collapsing volume.

We trained a U-Net architecture \cite{ronneberger2015u} to produce a dense flow vector field in the image space. U-Net has emerged as a mature technology particularly in the field of medical image segmentation. The U-Net model is able to produce an output that is at the original resolution of the image space. Further, the use of skip connections between the encoder and decoder arm equip the model with local and global context in the image. This makes the U-Net architecture ideally suited to our application since we wish to predict a dense flow vector field at the same resolution as the input image. Prior flow based methods \cite{lebrat2021corticalflow,bongratz2023meshes} have also successfully employed the U-Net in their model design.

Since the position of the template is no longer fixed but depends on the result of the linear transformation, the model requires a representation of the template mesh's current position. We achieve this by providing an unsigned distance map, $d(x)$, that encodes the template mesh's position in the image space, 

\begin{align}
    d(x) = \min_{y \in T} |x - y|_2 \label{eq:unsigned-distance-function}
\end{align}

\noindent where $T$ is the surface of the template mesh and $|\cdot|_2$ is the L2 distance norm. This distance map is concatenated to the image as a second channel and serves as the input to the U-Net. 

To avoid excessive compute at training time, we construct the distance map for the initial template ahead of time and simultaneously apply the same learned linear transformation to the template and the distance map. The distance map is computed from the center of each voxel to the nearest point on the surface of the initial template mesh. We use routines available in PyTorch3d to compute the distance to surface. We empirically observed that the distance map significantly improves the performance of the model. This approach can be an effective strategy to improve the performance of any multi-stage mesh deformation process. Namely, a distance function can effectively encode the location of the template mesh to standard computer vision models that expect inputs to be represented as dense arrays.

We performed numerical integration using an explicit 4th order Runge-Kutta scheme. Numerical integration is an approximation to the ``true'' integral and accumulates error over time. Prior flow-based approaches \cite{lebrat2021corticalflow} aim to minimize integration errors by controlling the time-step size based on the Lipschitz constant of the flow vector field. However, by doing this, the time-step size of integration is controlled by the region with the largest Lipschitz constant, possibly resulting in undesirably small time steps globally. We propose here an alternative strategy to mitigate the accumulation of integration error. We enforce the condition that a vertex cannot travel more than a distance of 1 voxel per time step by upper-bounding the L2 norm of the flow vector field. This condition is similar to the Courant-Friedrichs-Lewy (CFL) condition in numerical analysis. We refer the reader to a standard resource on numerical analysis such as \cite{leveque2007finite} for a more detailed discussion on the CFL condition. If $v$ is the predicted flow vector field before clipping, we compute the clipped flow vector field $v_{\mathrm{clip}}$ as,

\begin{align}
    v_{\mathrm{clip}} = \alpha \frac{v}{\max (|v|_2, \alpha)} \label{eq:clip-flow}
\end{align}

\noindent where $\alpha$ is a parameter that is set to approximately 1-voxel spacing. Notice that $v_{\mathrm{clip}}$ has an L2 norm upper-bounded by $\alpha$. Since the normalized image coordinates is a cube ranging across $[0,1]^3$ and is discretized into $128^3$ uniform voxels, the voxel-spacing in normalized image coordinates is $1/128 \approx 0.0078$. Thus we took $\alpha = 0.0075 < 1/128$. By clipping the flow-field, we are able to use a uniform step-size throughout the training process. We empirically observed that clipping the flow vector field resulted in a reduction in the number of self-intersecting faces suggesting that this simple strategy can be effective at controlling error due to numerical integration.

\begin{figure}[t]
    \centering
    \includegraphics[width=0.47\textwidth]{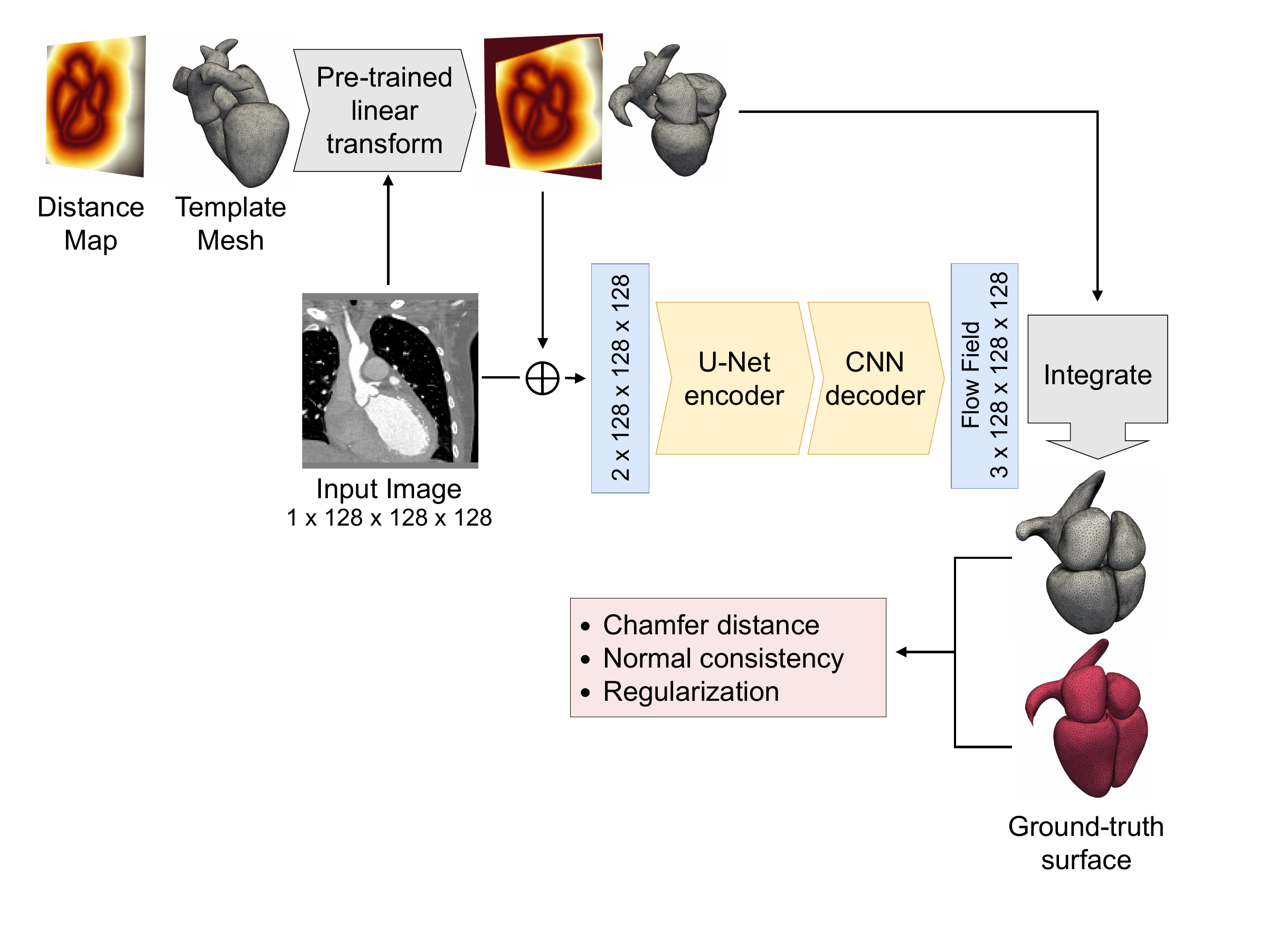}
    \caption{Workflow illustrating the training pipeline for the flow deformation module}
    \label{fig:flow-deformation}
\end{figure}

\subsection{Loss Functions}

The model is trained using a weighted sum of the following losses:

\begin{enumerate}
    \item Chamfer distance in the L1 norm as defined in eq. \ref{eq:chamfer}.
    \item Normal consistency between template and ground truth meshes.
    \item Volume loss as defined in eq. \ref{eq:volume-loss}.
    \item Mesh regularization, which includes edge length, normal consistency across faces, and Laplacian smoothing loss.
\end{enumerate}

\noindent Loss terms (1) and (4) are directly available in PyTorch3D \cite{ravi2020pytorch3d}.

\subsubsection*{Chamfer distance.} \label{sec:chamfer-distance}

The chamfer distance between two point clouds $P_1$ and $P_2$ can be computed as follows,

\begin{align}
    \mathrm{chamfer}(P_1, P_2) &= \frac{1}{P_1} \sum_{x \in P_1} \min_{y \in P_2} | x - y |_1 + \nonumber \\ 
    &\ \frac{1}{P_2} \sum_{y \in P_2} \min_{x \in P_1} | x - y|_1 \label{eq:chamfer}
\end{align}

Minimizing this loss leads to the model trying to increase the overlap between the point clouds. This has the effect of improving the accuracy of the model in terms of overlap with the grounds-truth geometry.

\subsubsection{Normal consistency between template and ground truth.} \label{sec:normal-consistency}

The normal consistency loss helps to associate surfaces with the correct orientation. This is typically achieved by computing the cosine similarity between the normal vectors from the template mesh with the normal vectors at the closest point in the ground-truth mesh and vice versa. This loss function is particularly important for the myocardium which is a cup-like structure. Since its inner and outer surfaces are separated by a small length scale, incorrectly associating these two surfaces will result in the two surfaces collapsing into each other. The normal consistency loss helps to prevent this. Note that the PyTorch3D implementation does not distinguish between the orientation of the two surfaces i.e. the sign of the normals does not affect the loss. This distinction is important for our application. For structures with thickness, not making this distinction can result in incorrectly associating surfaces together resulting in the collapse of this thickness. Instead, we use the following form of the normal consistency loss, which is a slight modification of the loss implemented in PyTorch3D. Given two point-sets $P, Q$ and a normal function $n(\cdot)$ which gives the normal at any point, the normal consistency loss $L_{NC}$ is,

\begin{align}
    L_{NC} (P,Q) &= \frac{1}{|P|} \sum_{x \in P} 1 - n(x)^T n(y) \nonumber \\
    \mathrm{s.t.} \quad y &= \argmin_{z \in Q} |x - z|_2 \label{eq:normal-consistency}
\end{align}

\noindent We compute the average of $L_{NC}(T,G)$ and $L_{NC}(G,T)$ as the final loss where $T$ and $G$ are respectively the template mesh and the ground truth mesh.

\subsubsection{Volume loss.} \label{sec:volume-loss}

To address the issue of collapsing volumes, we introduce a  physics-based loss function derived from the kinematics of continuous media. Consider a mesh vertex located at $x(t)$ where $t$ is the time-like parameter of integration. We can take $t \in [0,1]$ without loss of generality with $x(0)$ being the initial location of the mesh vertex and $x(1)$ being the location of the vertex after integration along the flow vector field $v$. Let $V(0)$ and $V(1)$ represent the volume of a small neighborhood around $x$ in the initial and final states. Since the flow vector field transforms the entire space, it also transforms the volume $V(0)$ to $V(1)$. A standard result in continuum mechanics (see chapter 3 in \cite{abeyaratne1998continuum}) relates the rate of change of $V$ to the divergence of the flow vector field,

\begin{align*}
    \frac{\mathrm{d} V}{\mathrm{d}t} = V \ \mathrm{div}(v)
\end{align*}

Integrating this quantity from time $t = 0$ to 1 we get,

\begin{align*}
L_{vol} := \frac{V_0}{V_1} = \mathrm{exp} \left( - \int_{0}^{1} \mathrm{div} (v) 
\ \mathrm{dt} \right)
\end{align*}

Suppose the flow vector field $v$ is such that a region with finite initial volume $V_0$ collapses to an infinitesimally small volume $V_1$ (i.e. $V_1 << V_0$) then $L_{vol}$ evaluates to a large value. We call $L_{vol}$ the volume loss and demonstrate that including this quantity during training helps to prevent surfaces separated by a small distance from collapsing into each other. $L_{vol}$ is computed by computing the integral in the above equation along the trajectory of every mesh vertex for which this loss is applied. Even though $v$ is taken to be a stationary vector field (i.e. it is not time-dependent), for a given mesh vertex $\mathrm{div}(v)$ varies along its integration trajectory. We compute $\mathrm{div}(v)$ using a central finite-difference scheme in the image space. Subsequently, we simultaneously integrate the positions and accumulate the above integral for all mesh vertices to which this loss is applied. 

Note that the exponential term in the above equation can cause this loss and its associated gradients to blow up which could cause instability in training. To avoid this, we clip the upper bound of the integral point-wise prior to computing the exponential. The equation below is our final expression for the Volume Loss,

\begin{align}
    L_{vol} = \mathrm{exp} \left( \mathrm{min} \left( 3, - \int_{0}^{1} \mathrm{div} (v) \ \mathrm{dt} \right) \right) \label{eq:volume-loss}
\end{align}

\noindent This simple strategy resulted in stable training in our experiments. Alternatively, gradient clipping available in machine learning libraries such as PyTorch may be directly employed.

\subsection{Dataset Information} \label{sec:dataset}

For training and testing, we use the same dataset as \cite{kong2021deep}. The training data consists of data from four public datasets including the multi-modality whole heart segmentation challenge (MMWHS) \cite{zhuang2019evaluation}, orCaScore challenge \cite{wolterink2016evaluation}, left atrial wall thickness challenge (SLAWT) \cite{karim2018algorithms}, and left-atrial segmentation challenge (LASC) \cite{tobon2015benchmark}. In total we had 101 CT samples and 47 MR samples in our dataset. We split this into a training dataset (86 CT, 41 MR) and a validation dataset (15 CT, 6 MR) for hyper-parameter tuning. The MMWHS challenge further provides a held-out test dataset consisting of 40 CT and 40 MR samples for which no ground-truth is available. Instead, the challenge organizers provide encrypted scripts that can be used to evaluate predicted segmentations on a variety of accuracy metrics. We use this held-out test dataset to evaluate the final performance of our models and report the same in the subsequent section. The input images are available in the NIfTI file format. Other image formats can be supported so long as the appropriate backend is available to load these files and convert them into image arrays. 

\begin{figure}[t]
    \centering
    \includegraphics[width=0.47\textwidth]{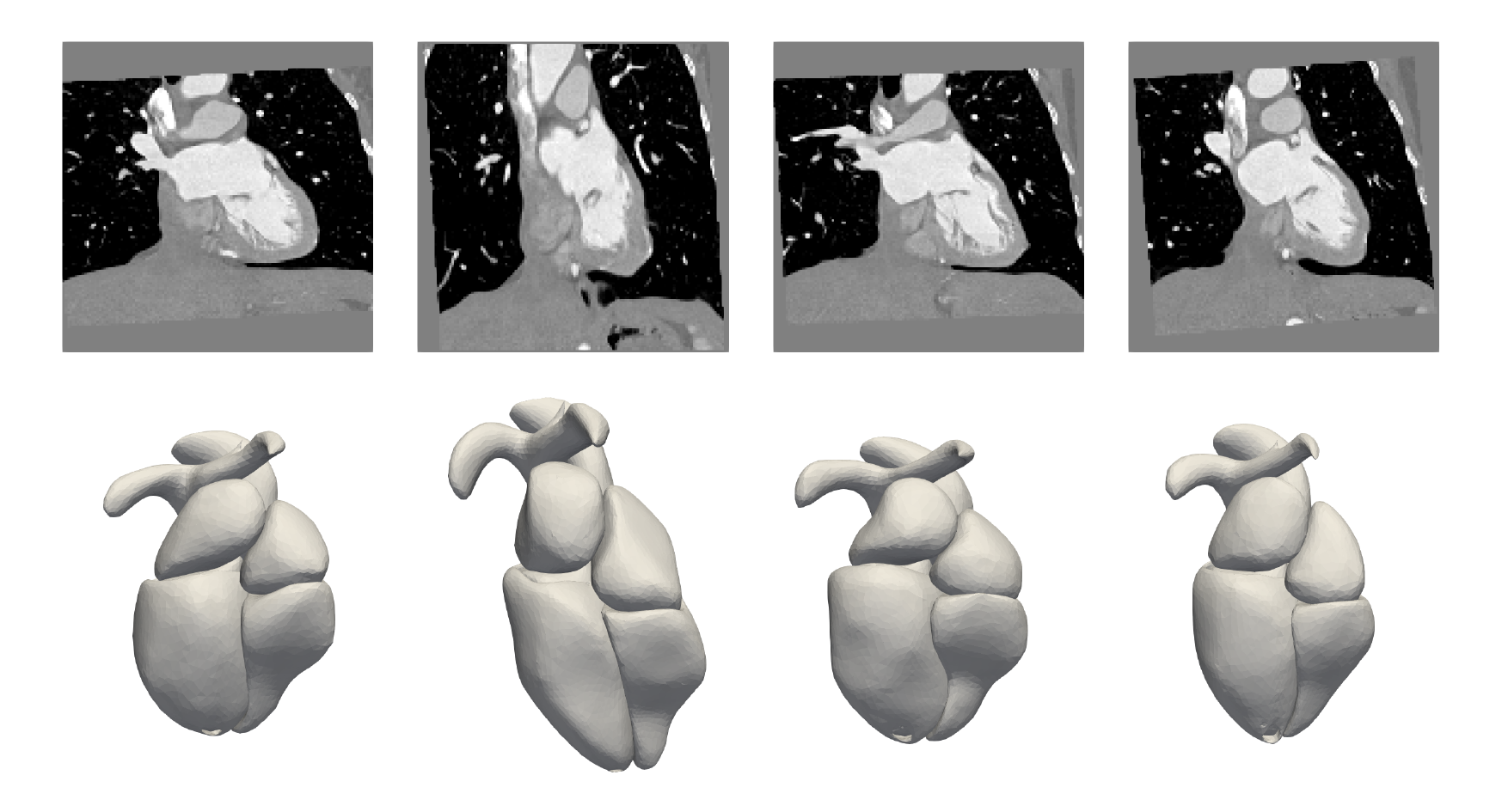}
    \caption{An illustration of samples generated by the data-augmentation process from a given input image and associated ground-truth segmentation. We generate these samples by applying small perturbations including random scaling, translation, rotation, shear, and local b-spline deformations to the input image and segmentation. Meshes are produced from the generated segmentations via the marching cubes algorithm.}
    \label{fig:data-augmentation}
\end{figure}

We augment the training dataset with small perturbations including random scaling, translation, rotation, shear, and local b-spline deformations. We produce 20 random augmentations per image. Figure \ref{fig:data-augmentation} shows typical image samples and associated ground-truth surfaces generated by the data-augmentation procedure for a single input image and segmentation. Subsequently, the dataset is pre-processed and normalized prior to training following a procedure similar to \cite{kong2020automating}. The input images are resampled to a standard dimension of $128^3$. Further, the voxel intensities are thresholded and normalized to lie between $[-1,1]$. We apply different normalization procedures to CT and MR samples since the Hounsfeld intensity values for CT images is standardized. For CT samples we threshold voxel intensities to lie between $[-750,750]$ and then scale the values to $[-1,1]$. We threshold MR voxel intensities at the 20\textsuperscript{th} percentile of values (lower) and 99\textsuperscript{th} percentile of values (upper) and scale the resulting values to $[-1,1]$. We apply thresholding to ensure that the voxel intensities in the range expressed by the cardiac structures of interest are captured, while largely removing intensity variations due to bones or imaging artefacts. Ground-truth meshes were generated using the marching cube algorithm on the ground-truth segmentations, followed by mesh smoothing.

Since the original MR volumes had a significantly larger field-of-view, we cropped the images such that the cardiac structures occupied a similar proportion of the image space as in the CT data. This was automated in the training data by using the ground truth segmentations to determine the cropping dimensions. The test data was manually cropped, however automatic methods to generate a bounding box \cite{payer2017multi} may be used in the future to avoid manual cropping. We evaluated our models on the held-out test dataset of the MMWHS Challenge \cite{zhuang2019evaluation}.

\section{Results} \label{sec:results}

We trained a single model on both CT and MR data using the hyperparameters listed in table \ref{tab:hyperparameters}. This corresponds to model LT-FL-V. For model LT-FL we set the weight of the volume loss to zero. We used a learning rate scheduler to reduce the learning rate when the validation loss plateaus for a fixed number of iterations. The model was trained on a single Nvidia 2080TI GPU for about 24 hours. 

\begin{table*}[t]
\caption{Hyperparameter settings used to train the model}
    \begin{center}
    \label{tab:hyperparameters}
    \begin{tabular}{| c | c |} \hline
Hyperparameter & Value \\ \hline
Batch size & 1 \\
Initial learning rate & $5 \times 10^{-5}$ \\
Flow norm threshold & 0.0075 \\ \hline
\end{tabular}
\begin{tabular}{| c | c |} \hline
Loss Function & Weight \\ \hline
Chamfer distance & 1 \\
Chamfer normal consistency & 0.20 \\
Volume loss & 0.005 \\
Edge loss & 50 \\
Normal consistency loss & 1 \\
Laplace smoothing loss & 30 \\ \hline
\end{tabular}
\end{center}
\end{table*}

\subsection{Model Accuracy} \label{sec:model-accuracy}

To measure our model's accuracy, we converted our generated meshes into segmentations. The VTK \cite{vtkBook} library provides routines to convert a mesh into an image mask. We used these routines to obtain an image mask consisting of the region enclosed by the mesh in the image space. We obtained one mask per cardiac structure and consider these as the segmentations corresponding to our predicted meshes. We then evaluated the accuracy of these segmentations using the scripts provided by the MMWHS challenge organizers. We consider HeartDeformNet \cite{Kong2022LearningWH} as the benchmark for this problem and compare our accuracy and mesh quality against the results reported in that work. Table \ref{tab:ct-accuracy-comparison} compares accuracy metrics including Dice Score, Jaccard Index, Average Symmetric Surface Distance (ASSD) and Hausdorff Distance (HD). Similarly Table \ref{tab:mr-accuracy-comparison} compares these accuracy metrics for the MR test dataset. The reported results are the average over the 40 test samples in each dataset with the standard deviation reported in parantheses.

We observe that the performance of our model is commensurate similar with HeartDeformNet on average being slightly more accurate. We note that HeartDeformNet deforms the mesh over 3 successive stages each of which is a graph convolutional neural network operating on the mesh. We anticipate that by incorporating a second deformation block, or fine-tuning a pre-trained network, we can substantially improve our performance.

\subsection{Mesh Quality} \label{sec:mesh-quality}

Next we compare the quality of the generated meshes in terms of the percentage self-intersecting faces (SIF). We measure SIF using the PyMesh library \cite{zhou2020pymesh}. SIF measures intersections due to surface inter-penetration and intersections due to element inversion. The usefulness of our method is clearly seen in Table \ref{tab:ct-sif-comparison} and Table \ref{tab:mr-sif-comparison}. These tables report the percentage SIF for each cardiac structure for CT and MR respectively measured on the meshes generated from the test dataset.  Notice that for CT samples each of the 40 generated meshes for the CT test dataset had \emph{zero} self-intersecting faces across all cardiac structures. Similarly, 38 out of 40 generated meshes for the MR test dataset showed \emph{zero} self-intersecting faces across all cardiac structures. Of the two samples that had non-zero SIF, one had 0.85\% SIF in the Aorta and another sample had 0.92\% SIF in the pulmonary artery. We note that our template mesh had a higher resolution compared to the template used by HeartDeformNet. Our template mesh consists of 110K triangular faces and the template used by HeartDeformNet consists of 45K faces. Generally speaking, higher resolution meshes are more prone to self-intersections due to element collapse. Despite having a significantly higher number of elements, our model is able to consistently generate meshes with far fewer self-intersecting faces.

The ability to reliably generate meshes with zero self-intersections is a very useful property because the presence of even a few self-intersecting faces can cause a mesh generation software like TetGen \cite{hang2015tetgen} to abort. Thus, our model makes significant progress towards robustly producing simulation-ready meshes that do not require any post-processing before being usable in a simulation environment.


\begin{table*}[t]
\footnotesize
\caption{Comparing mean (standard deviation) of accuracy metrics on CT test data. $\uparrow (\downarrow$) indicates higher (lower) value is better.}
    \begin{center}
    \label{tab:ct-accuracy-comparison}
    \begin{tabular}{| c | c | c | c | c | c | c | c | c | c |} \hline
    Metric  & Model          & Epi          & LA           & LV           & RA           & RV           & Ao          & PA          & WH            \\ \hline
Dice ($\uparrow$)    & Ours           &  0.86 (0.04) &  0.92 (0.03) &  0.91 (0.06) &  0.85 (0.08) &  0.88 (0.04) & 0.89 (0.04) & 0.82 (0.09) &  0.89 (0.03)  \\ 
        & HeartDeformNet &  0.88 (0.03) &  0.93 (0.03) &  0.92 (0.04) &  0.89 (0.05) &  0.91 (0.03) & 0.91 (0.04) & 0.85 (0.09) &  0.91 (0.02)  \\ \hline
Jaccard ($\uparrow$) & Ours           &  0.76 (0.06) &  0.85 (0.05) &  0.84 (0.09) &  0.75 (0.11) &  0.79 (0.07) & 0.81 (0.06) & 0.70 (0.12) &  0.80 (0.04)  \\ 
        & HeartDeformNet &  0.79 (0.05) &  0.86 (0.05) &  0.85 (0.06) &  0.80 (0.07) &  0.83 (0.05) & 0.84 (0.06) & 0.74 (0.13) &  0.83 (0.04)  \\ \hline
ASSD ($\downarrow$)  & Ours           &  1.40 (0.36) &  1.20 (0.33) &  1.16 (0.47) &  1.82 (0.72) &  1.34 (0.39) & 1.13 (0.35) & 1.53 (0.75) &  1.36 (0.26)  \\ 
        & HeartDeformNet &  1.38 (0.24) &  1.14 (0.38) &  1.10 (0.33) &  1.63 (0.72) &  1.14 (0.27) & 0.93 (0.38) & 1.20 (0.74) &  1.25 (0.24)  \\ \hline
HD  ($\downarrow$)    & Ours           & 13.74 (2.97) & 12.21 (2.20) & 12.55 (2.05) & 12.52 (4.83) & 10.80 (4.74) & 5.79 (1.70) & 7.84 (3.01) & 16.32 (4.00)  \\ 
        & HeartDeformNet & 14.40 (2.75) &  8.18 (3.08) &  6.87 (2.51) & 12.46 (5.99) &  9.55 (2.01) & 5.54 (1.91) & 8.45 (2.96) & 16.63 (4.37)  \\ \hline
    \end{tabular}
    \end{center}
\end{table*}

\begin{table*}[t]
\footnotesize
\caption{Comparing mean (standard deviation) of accuracy metrics on MR test data. $\uparrow (\downarrow$) indicates higher (lower) value is better.}
    \begin{center}
    \label{tab:mr-accuracy-comparison}
    \begin{tabular}{| c | c | c | c | c | c | c | c | c | c |} \hline
Matric  & Model           &  Epi           &   LA          &   LV           &  RA          &  RV          & Ao           &  PA          & WH            \\ \hline  
Dice ($\uparrow$)   & Ours            &   0.76 (0.09)  &   0.85 (0.05) &   0.90 (0.03)  &  0.86 (0.04) &  0.85 (0.09) & 0.84 (0.06)  &  0.69 (0.14) &  0.84 (0.05)  \\ 
        & HeartDeformNet  &   0.79 (0.10)  &   0.86 (0.08) &   0.89 (0.06)  &  0.88 (0.04) &  0.87 (0.06) & 0.83 (0.07)  &  0.78 (0.12) &  0.86 (0.05)  \\ \hline
Jaccard ($\uparrow$) & Ours            &   0.62 (0.11)  &   0.74 (0.08) &   0.82 (0.05)  &  0.76 (0.06) &  0.74 (0.12) & 0.73 (0.08)  &  0.54 (0.15) &  0.73 (0.07)  \\ 
        & HeartDeformNet  &   0.66 (0.11)  &   0.77 (0.10) &   0.81 (0.09)  &  0.78 (0.06) &  0.78 (0.09) & 0.72 (0.10)  &  0.65 (0.14) &  0.76 (0.07)  \\ \hline
ASSD ($\downarrow$) & Ours            &   2.22 (1.14)  &   1.72 (0.67) &   1.46 (0.51)  &  1.79 (0.44) &  1.97 (1.36) & 1.48 (0.54)  &  2.37 (0.92) &  1.86 (0.69)  \\ 
        & HeartDeformNet  &   2.10 (1.24)  &   1.57 (0.66) &   1.54 (0.77)  &  1.58 (0.49) &  1.56 (0.72) & 1.53 (0.66)  &  1.66 (0.73) &  1.66 (0.59)  \\ \hline
HD  ($\downarrow$)  & Ours            &  16.90 (3.48)  &  12.76 (3.30) &  13.86 (6.23)  & 11.80 (3.15) & 12.99 (9.69) & 7.71 (4.72)  & 10.29 (3.70) & 20.59 (8.96)  \\ 
        & HeartDeformNet  &  15.96 (3.33)  &  10.16 (3.77) &   8.97 (6.74)  & 12.56 (4.51) & 12.46 (9.12) & 7.39 (3.39)  &  9.14 (3.43) & 18.91 (9.24)  \\ \hline

    \end{tabular}
    \end{center}
\end{table*}

\begin{table*}[t]
\caption{Comparing mean (max) percentage self-intersecting faces on CT test dataset}
    \begin{center}
    \label{tab:ct-sif-comparison}
    \begin{tabular}{| c | c | c | c | c | c | c | c | c |} \hline
Model            &   Epi	    &   LA      &    LV     &	RA          &	RV     &   Ao      &   PA      \\ \hline
Ours             &   0 (0)    &   0 (0)   &    0 (0)  &   0 (0)       &   0 (0)  &   0 (0)   &   0 (0)       \\ 
HeartDeformNet	&   0 (0)	 &   0 (0)   &    0	(0)  &   0.01 (0.25) &   0 (0)	&   0 (0)	&   0.18 (1.14)   \\ \hline
    \end{tabular}
    \end{center}
\end{table*}

\begin{table*}[t]
\caption{Comparing mean (max) percentage self-intersecting faces on MR test dataset}
    \begin{center}
    \label{tab:mr-sif-comparison}
    \begin{tabular}{| c | c | c | c | c | c | c | c | c |} \hline
Model &  Epi          &  LA    &  LV           &  RA          &  RV           &  Ao           &  PA           \\ 	\hline
Ours &  0 (0)        &  0 (0) &  0 (0)        &  0 (0)       &  0 (0)        &  0.02 (0.85)  &  0.023 (0.92) \\	
HeartDeformNet &  0.019 (0.32) &  0 (0) &  0.013 (0.38) &  0.03 (0.07) &  0.012 (0.31) &  0.138 (2.96) &  0.625 (4.76) \\	\hline
    \end{tabular}
    \end{center}
\end{table*}

\subsection{Ablation Study of Model Architecture} \label{sec:architecture-ablation}

We investigated the different elements of our proposed model architecture and compared the accuracy and quality of the generated meshes. We compare the following model design choices here,

\begin{enumerate}
    \item LT - A model that \emph{only} employs the linear transformation.
    \item FL - A model that \emph{only} employs the flow deformation \emph{without} the linear transformation.
    \item LT-FL - A model combining the linear transformation and flow deformation but trained without the volume loss eq. \ref{eq:volume-loss}.
    \item LT-FL-v - The LT-FL model that is trained additionally with the volume loss eq. \ref{eq:volume-loss}.
\end{enumerate}

\begin{figure}[t]
    \centering
    \includegraphics[width=0.5\textwidth]{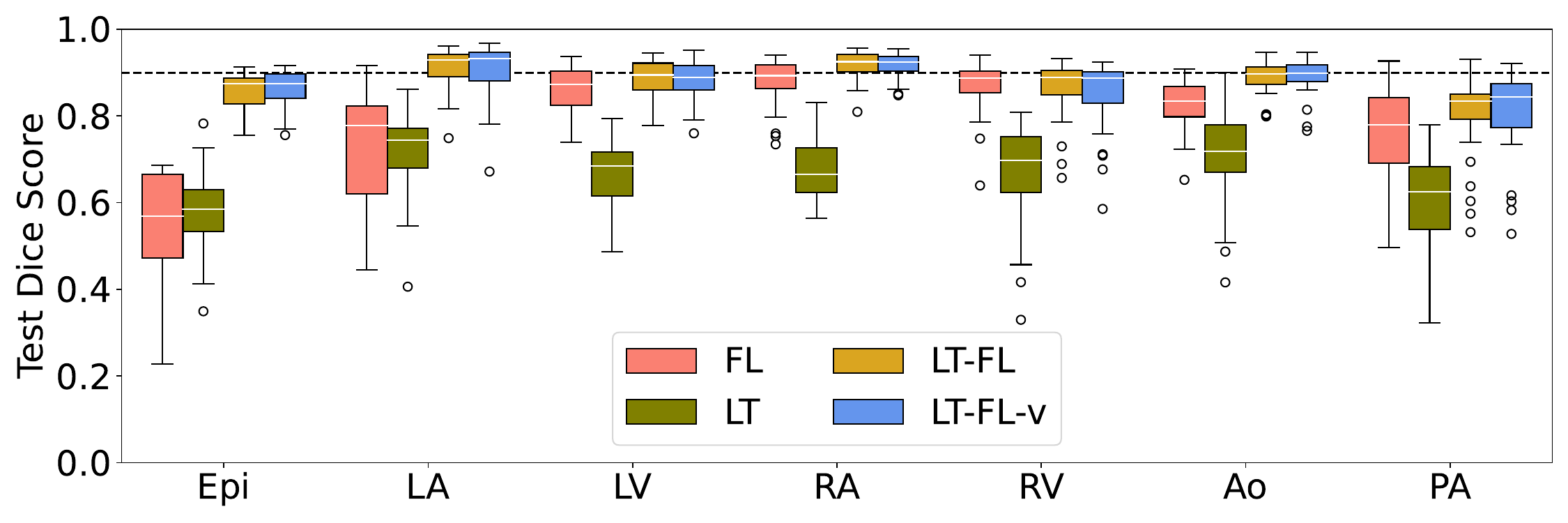}
    \caption{Comparison of dice score on CT test data for different model choices}
    \label{fig:dice-ablation-ct}
\end{figure}

\begin{figure}[t]
    \centering
    \includegraphics[width=0.5\textwidth]{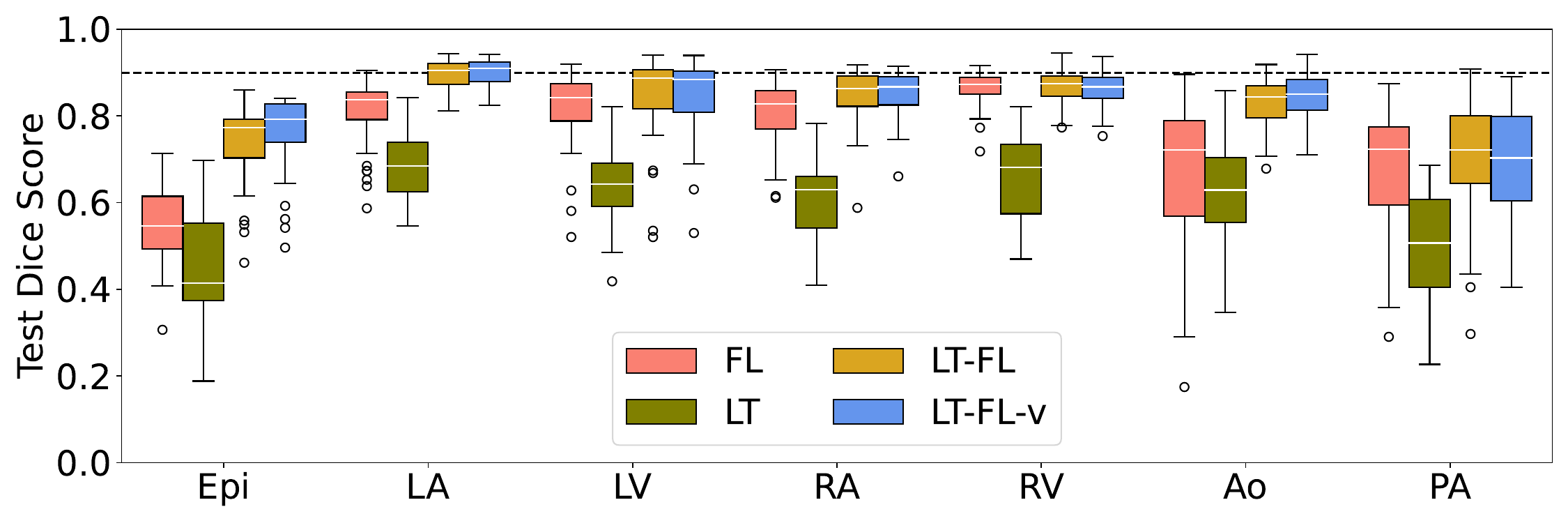}
    \caption{Comparison of dice score on MR test data for different model choices}
    \label{fig:dice-ablation-mr}
\end{figure}

To assess the performance of these models, we generated meshes on the held-out test dataset. Figures \Ref{fig:dice-ablation-ct} and \ref{fig:dice-ablation-mr} compare the dice scores of the different models. Detailed comparison of all accuracy metrics are provided in tables \ref{tab:ct-ablation-accuracy} and \ref{tab:mr-ablation-accuracy}. We observed that combining the linear transformation with the flow deformation substantially improves the accuracy and quality of the generated meshes in almost all cases. As expected, the linear transformation introduces zero self-intersections in all cases, which benefits the LT-FL model as well. Further, the volume loss has a powerful regularization effect on the generated flow fields, which helps to almost completely eliminate self-intersections.

\begin{table*}[t]
\caption{Ablation study -- comparing mean (standard deviation) of accuracy metrics on CT test data. $\uparrow (\downarrow$) indicates higher (lower) value is better.}
\footnotesize
    \begin{center}
    \label{tab:ct-ablation-accuracy}
    \begin{tabular}{| c | c | c | c | c | c | c | c | c | c |} \hline
Metric  & Model		& Epi		   &	LA			 &	LV			 &	RA			 &	RV			 &	Ao			 &	PA			 &	WH			 \\ \hline
Dice ($\uparrow$)	& Fl			&  0.56 (0.12) &	 0.88 (0.05) &	 0.73 (0.14) &	 0.87 (0.06) &	 0.86 (0.05) &	 0.83 (0.06) &	 0.76 (0.10) &	 0.79 (0.05) \\ 
		& LT			&  0.58 (0.09) &	 0.68 (0.07) &	 0.72 (0.09) &	 0.67 (0.11) &	 0.66 (0.08) &	 0.71 (0.11) &	 0.61 (0.11) &	 0.66 (0.06) \\ 
		& LT-Fl		&  0.86 (0.04) &	 0.92 (0.03) &	 0.91 (0.05) &	 0.87 (0.06) &	 0.89 (0.04) &	 0.89 (0.03) &	 0.81 (0.09) &	 0.89 (0.02) \\ 
		& LT-Fl-v		&  0.86 (0.04) &	 0.92 (0.03) &	 0.91 (0.06) &	 0.85 (0.08) &	 0.88 (0.04) &	 0.89 (0.04) &	 0.81 (0.09) &	 0.89 (0.03) \\ \hline
Jaccard ($\uparrow$)	& Fl			&  0.39 (0.11) & 	 0.79 (0.08) &	 0.59 (0.16) &	 0.77 (0.09) &	 0.76 (0.08) &	 0.71 (0.08) &	 0.63 (0.13) &	 0.66 (0.07) \\ 
		& LT			&  0.41 (0.09) &	 0.51 (0.08) &	 0.57 (0.10) &	 0.51 (0.12) &	 0.50 (0.09) &	 0.56 (0.13) &	 0.44 (0.11) &	 0.49 (0.06) \\ 
		& LT-Fl		&  0.75 (0.06) &	 0.85 (0.05) &	 0.84 (0.07) &	 0.77 (0.09) &	 0.80 (0.06) &	 0.80 (0.05) &	 0.69 (0.12) &	 0.80 (0.04) \\ 
		& LT-Fl-v		&  0.76 (0.06) &	 0.85 (0.05) &	 0.84 (0.09) &	 0.75 (0.11) &	 0.79 (0.07) &	 0.81 (0.06) &	 0.69 (0.12) &	 0.80 (0.04) \\ \hline
ASSD ($\downarrow$)	& Fl			&  1.97 (1.10) &	 1.54 (0.41) &	 4.40 (1.85) &	 1.86 (0.85) &	 1.75 (0.51) &	 2.02 (0.84) &	 2.00 (0.89) &	 2.24 (0.52) \\ 
		& LT			&  3.63 (0.97) &	 4.28 (1.05) &	 4.01 (1.04) &	 4.44 (1.21) &	 4.59 (1.05) &	 2.94 (1.09) &	 3.71 (1.23) &	 4.07 (0.64) \\ 
		& LT-Fl		&  1.33 (0.38) &	 1.22 (0.36) &	 1.16 (0.37) &	 1.70 (0.69) &	 1.36 (0.43) &	 1.17 (0.30) &	 1.64 (0.77) &	 1.35 (0.24) \\ 
		& LT-Fl-v		&  1.43 (0.36) &	 1.25 (0.33) &	 1.17 (0.47) &	 1.82 (0.72) &	 1.33 (0.39) &	 1.13 (0.35) &	 1.55 (0.75) &	 1.37 (0.26) \\ \hline
HD	($\downarrow$)	& Fl			& 13.90 (9.09) &	10.81 (3.13) &	12.68 (2.84) &	12.01 (5.26) &	12.10 (5.11) &	10.40 (5.37) &	10.61 (4.56) &	18.86 (9.09) \\ 
		& LT			& 16.80 (2.46) &	14.75 (3.47) &	14.88 (3.37) &	17.82 (4.38) &	16.71 (3.30) &	 9.79 (2.75) &	12.69 (3.99) &	20.34 (3.80) \\ 
		& LT-Fl		& 12.59 (2.95) &	 9.79 (3.08) &	 8.88 (3.08) &	11.73 (4.99) &	10.65 (4.71) &	 5.96 (1.82) &	 8.08 (2.92) &	15.86 (4.19) \\ 
		& LT-Fl-v		& 13.60 (2.97) &	12.15 (2.20) &	12.30 (2.05) &	12.55 (4.83) &	11.28 (4.74) &	 5.74 (1.70) &	 7.92 (3.01) &	16.21 (4.00) \\ \hline
    \end{tabular}
    \end{center}
\end{table*}

\begin{table*}[t]
\footnotesize
\caption{Ablation study -- comparing mean (standard deviation) of accuracy metrics on MR test data. $\uparrow (\downarrow$) indicates higher (lower) value is better.}
    \begin{center}
    \label{tab:mr-ablation-accuracy}
    \begin{tabular}{| c | c | c | c | c | c | c | c | c | c |} \hline
Metric	& Model	& Epi			& 	LA			&	LV			&	RA			&	RV			&	Ao			&	PA			& 	WH		   \\ \hline
Dice ($\uparrow$)	& Fl		& 0.55	(0.09)	& 0.81	(0.07)	& 0.81	(0.08)	& 0.86	(0.04)	& 0.82	(0.09)	& 0.67	(0.18)	& 0.68	(0.15)	& 0.77	(0.05) \\ 
		& LT		& 0.45	(0.12)	& 0.62	(0.09)	& 0.69	(0.08)	& 0.66	(0.10)	& 0.64	(0.08)	& 0.62	(0.12)	& 0.50	(0.13)	& 0.60	(0.06) \\ 
		& LT-Fl	& 0.74	(0.09)	& 0.85	(0.06)	& 0.89	(0.04)	& 0.87	(0.04)	& 0.85	(0.10)	& 0.83	(0.06)	& 0.70	(0.14)	& 0.84	(0.05) \\ 
		& LT-Fl-v	& 0.76	(0.09)	& 0.85	(0.05)	& 0.90	(0.03)	& 0.86	(0.04)	& 0.85	(0.09)	& 0.84	(0.06)	& 0.69	(0.14)	& 0.84	(0.05) \\ \hline
Jaccard ($\uparrow$)	& Fl		& 0.38	(0.08)	& 0.68	(0.10)	& 0.69	(0.10)	& 0.76	(0.06)	& 0.70	(0.12)	& 0.52	(0.19)	& 0.53	(0.16)	& 0.63	(0.07) \\ 
		& LT		& 0.30	(0.11)	& 0.45	(0.09)	& 0.53	(0.09)	& 0.51	(0.11)	& 0.48	(0.09)	& 0.46	(0.12)	& 0.34	(0.11)	& 0.43	(0.06) \\ 
		& LT-Fl	& 0.60	(0.11)	& 0.74	(0.09)	& 0.81	(0.06)	& 0.77	(0.06)	& 0.74	(0.13)	& 0.71	(0.08)	& 0.55	(0.16)	& 0.73	(0.07) \\ 
		& LT-Fl-v	& 0.62	(0.11)	& 0.74	(0.08)	& 0.81	(0.05)	& 0.76	(0.06)	& 0.75	(0.12)	& 0.72	(0.08)	& 0.55	(0.15)	& 0.73	(0.07) \\ \hline
ASSD ($\downarrow$)	& Fl		& 2.89	(1.18)	& 2.00	(0.94)	& 3.00	(1.09)	& 1.89	(0.54)	& 2.20	(0.86)	& 3.14	(1.51)	& 2.46	(1.03)	& 2.49	(0.69) \\ 
		& LT		& 5.20	(1.43)	& 4.47	(1.37)	& 4.94	(1.55)	& 4.85	(1.96)	& 5.20	(1.53)	& 3.60	(1.25)	& 4.61	(1.56)	& 4.97	(0.82) \\ 
		& LT-Fl	& 2.46	(1.16)	& 1.70	(0.69)	& 1.51	(0.60)	& 1.65	(0.47)	& 1.92	(1.30)	& 1.62	(0.48)	& 2.26	(0.94)	& 1.88	(0.68) \\ 
		& LT-Fl-v	& 2.26	(1.14)	& 1.72	(0.67)	& 1.48	(0.51)	& 1.78	(0.44)	& 1.96	(1.36)	& 1.52	(0.54)	& 2.34	(0.92)	& 1.86	(0.69) \\ \hline
HD	($\downarrow$)	& Fl		& 20.69	(7.37)	& 11.98	(4.69)	& 14.05	(6.35)	& 12.67	(4.20)	& 15.36	(9.26)	& 15.14	(6.16)	& 11.59	(3.95)	& 23.63	(9.94) \\ 
		& LT		& 21.23	(5.19)	& 17.04	(4.26)	& 20.55	(7.41)	& 18.65	(6.01)	& 22.83	(8.65)	& 12.66	(4.20)	& 14.94	(3.87)	& 27.25	(8.00) \\ 
		& LT-Fl	& 16.91	(4.52)	& 10.10	(3.53)	& 10.14	(7.23)	& 12.48	(4.51)	& 13.53	(9.82)	& 8.23	(4.45)	& 10.89	(4.57)	& 20.85	(9.29) \\ 
		& LT-Fl-v	& 16.88	(3.48)	& 13.04	(3.30)	& 13.90	(6.23)	& 11.71	(3.15)	& 12.71	(9.69)	& 7.62	(4.72)	& 10.19	(3.70)	& 20.57	(8.96) \\ \hline
    \end{tabular}
    \end{center}
\end{table*}

\begin{table*}[t]
\caption{Ablation study -- comparing mean (max) percentage self-intersecting faces on CT test data}
    \begin{center}
    \label{tab:ct-ablation-quality}
    \begin{tabular}{| c | c | c | c | c | c | c | c | c |} \hline
Model	& Epi				& LA	& 	LV			& 	RA		    & 		RV		&		Ao			& 	PA			\\ \hline
Fl		& 0.07	(0.69)	& 0 (0)	& 0.007	(0.20)	& 0	  (0)		& 0.17	(1.67)	&	0.36	(10.16)	& 4.97	(18.12)	\\ 
LT		& 0		(0)    	& 0 (0)	& 0		(0)		& 0	  (0)		& 0		(0)		&	0		(0)		& 0		(0)		\\ 
LT-Fl	& 0.0005	(0.01)	& 0 (0)	& 0		(0)		& 0.014 (0.58)	& 0		(0)		&	0.19	(2.64)	& 0.20	(5.41)	\\ 
LT-Fl-v	& 0		(0)    	& 0 (0)	& 0		(0)		& 0	  (0)		& 0		(0)		&	0		(0)		& 0		(0)		\\ \hline
    \end{tabular}
    \end{center}
\end{table*}

\begin{table*}[t]
\caption{Ablation study -- comparing mean (max) percentage self-intersecting faces on MR test data}
    \begin{center}
    \label{tab:mr-ablation-quality}
    \begin{tabular}{| c | c | c | c | c | c | c | c | c |} \hline
Model	& Epi			& LA			& LV			& RA			& RV			& Ao				& PA			\\ \hline
Fl		& 0.06	(1.58)	& 0.05	(1.35)	& 0.004	(0.14)	& 0.04	(0.49)	& 0.06	(1.95)	& 0.06		(1.78)	& 4.14	(21.8)	\\ 
LT		& 0		(0)		& 0		(0)		& 0		(0)		& 0		(0)		& 0		(0)		& 0			(0)		& 0		(0)		\\ 
LT-Fl	& 0.002	(0.06)	& 0.009	(0.37)	& 0		(0)		& 0.03	(1.12)	& 0.001	(0.04)	& 0.69		(4.89)	& 0.09	(0.82)	\\ 
LT-Fl-v	& 0		(0)		& 0		(0)		& 0		(0)		& 0		(0)		& 0		(0)		& 0.0008	(0.85)	& 0.02	(0.92)	\\ \hline
    \end{tabular}
    \end{center}
\end{table*}

\subsection{Patient Specific Meshes of Left Ventricle With Tissue Thickness}

\begin{figure}[t]
    \centering
    \includegraphics[width=0.47\textwidth]{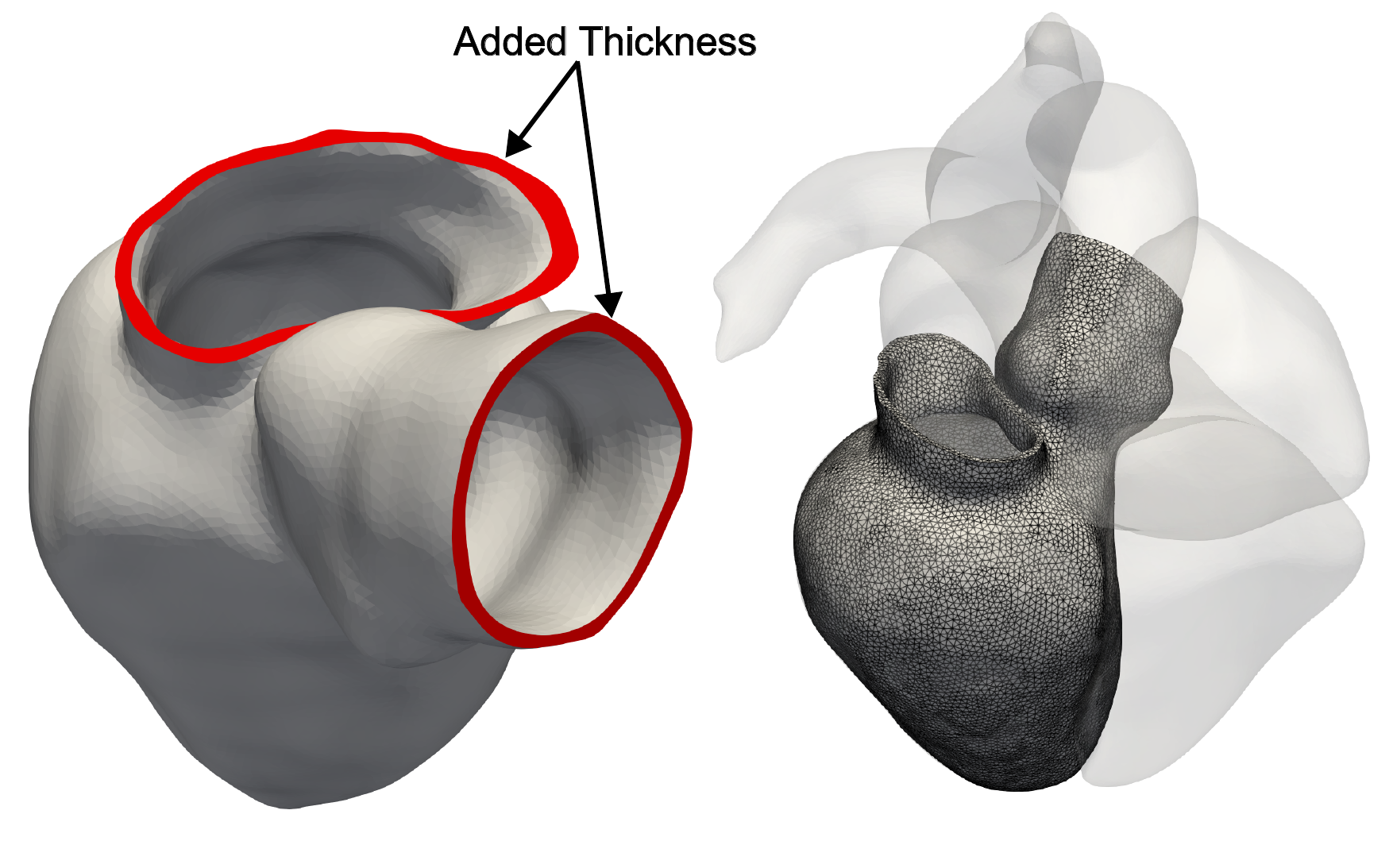}
    \caption{Template mesh of the left-ventricle myocardium with thickness added to inlets and outlets}
    \label{fig:lv-template}
\end{figure}

We now showcase an application of our method in generating meshes of cardiac structures wherein thickness information is not available in the ground truth. Figure \Ref{fig:lv-template} shows a template mesh of the left-ventricle myocardium that was generated by combining meshes of the myocardium, aorta, and left atrium. As shown in fig. \ref{fig:tissue-thickness}, the thickness of the aorta and left atrium is not visible in the original image and must be added based on knowledge of cardiac tissue.

It is important to note that since the ground truth surfaces of the aorta and left atrium do not contain thickness, the vector fields describing the deformation of the template in a neighborhood around these surfaces point towards these surfaces. In other words, these surfaces act like sinks in the deformation field. The model thus learns to collapse any wall thicknesses added in the template. This is undesirable for the following reasons:

\begin{enumerate}
    \item We will not be able to generate a volumetric mesh of the myocardium from these collapsed surfaces. Repairing these collapsed surfaces is non-trivial and requires significant manual effort. Further, the mechanical response of a collapsed surface will vastly over-exaggerate the true deformation of these tissues when subjected to mechanical load.
    \item We require the red surfaces in fig. \ref{fig:lv-template} to apply boundary conditions for electromechanics simulation of the LV. If these surfaces are collapsed, we will not be able to apply the necessary boundary conditions.
\end{enumerate}

A significant advantage of our method is that it is agnostic to the resolution of the template mesh and may be applied to any subset of the original template used for training. This enables us to deploy our model on the new template shown in fig. \ref{fig:lv-template} even though this template mesh is different from the whole heart template mesh used in training our model.

\begin{figure}[t]
    \centering
    \includegraphics[width=0.47\textwidth]{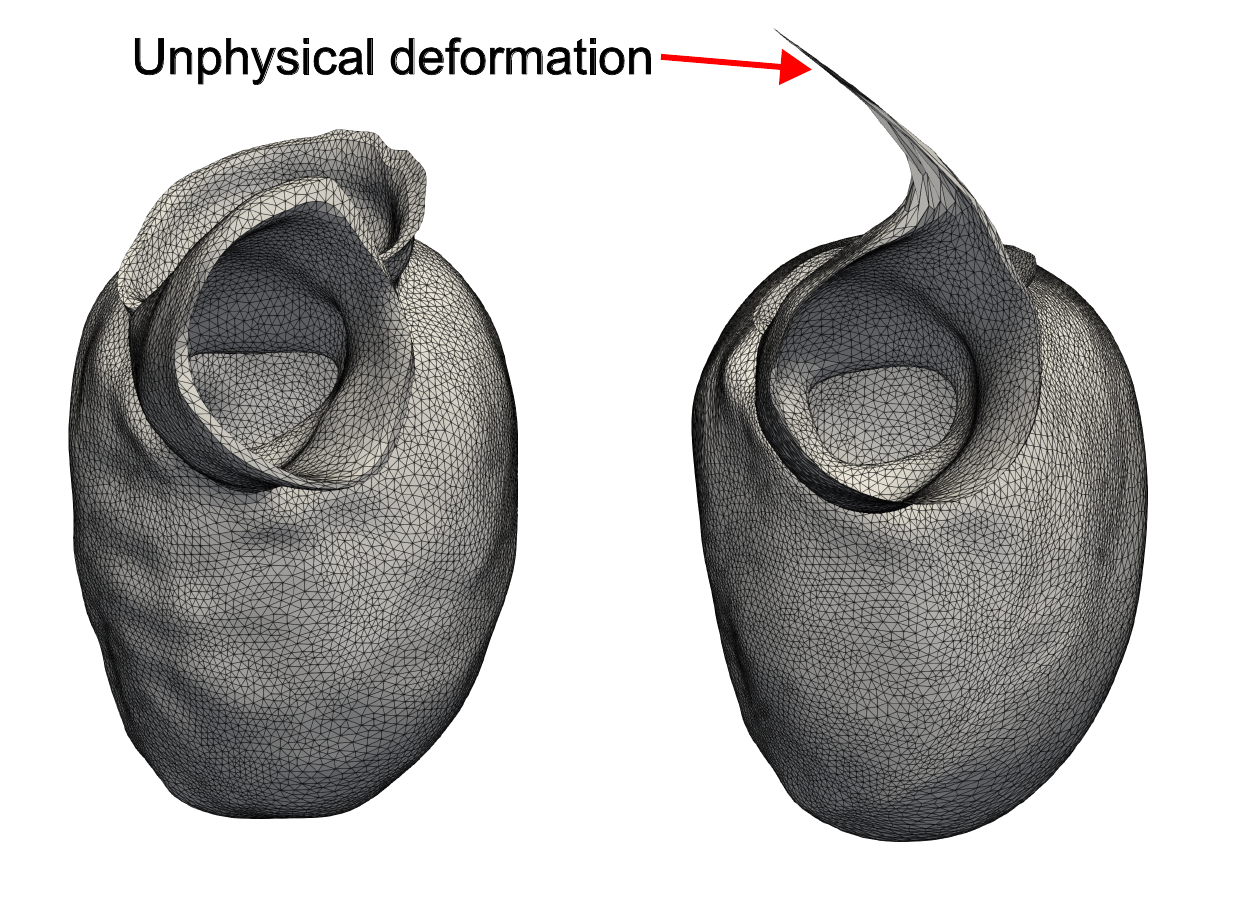}
    \caption{Volume loss is effective at regularizing unphysical deformations}
    \label{fig:lv-distortion}
\end{figure}

Table \Ref{tab:lv-mesh-sif} shows a comparison of SIF \% for the LT-FL and LT-FL-v models. Recall that the only difference between these two models is that the LT-FL-v additionally uses the volume loss eq. \ref{eq:volume-loss} during training. The LT-FL-v model on average reduces the number of SIF by over 90\% for CT and 93\% for MR compared to LT-FL with the median improvement being 100\% for both modalities. We see that more than half the samples for CT (50\%) and MR (62.5\%) have \emph{zero} SIF. This is clear evidence of the regularization effect provided by the volume loss and additionally helps prevent unphysical deformations as shown in fig. \ref{fig:lv-distortion}.

\begin{table*}[t]
\caption{LV-template mesh quality on CT and MR test data}
    \begin{center}
    \label{tab:lv-mesh-sif}
    \begin{tabular}{| c | c | c | c | c | } \hline
Modality & Metric          & LT-FL & LT-FL-v & HeartDeformNet \\ \hline
  CT     & Avg. SIF (\%)   & 0.109 & 0.003   & 0.134         \\
         & Median SIF (\%) & 0.078 & 0.002   & 0.112         \\
         & Max SIF (\%)    & 0.731 & 0.013   & 0.54          \\
         & \% Samples with 0\% SIF & 0 & 50  & 0             \\ \hline
MR       & Avg. SIF (\%)   & 0.138 & 0.002   & 0.145         \\
         & Median SIF (\%) & 0.080 & 0       & 0.086         \\
         & Max SIF (\%)    & 0.606 & 0.013   & 0.885         \\
         & \% Samples with 0\% SIF &  2.5 & 62.5 & 0 \\ \hline
    \end{tabular}
    \end{center}
\end{table*}

\begin{figure}[t]
    \centering
    \includegraphics[width=0.47\textwidth]{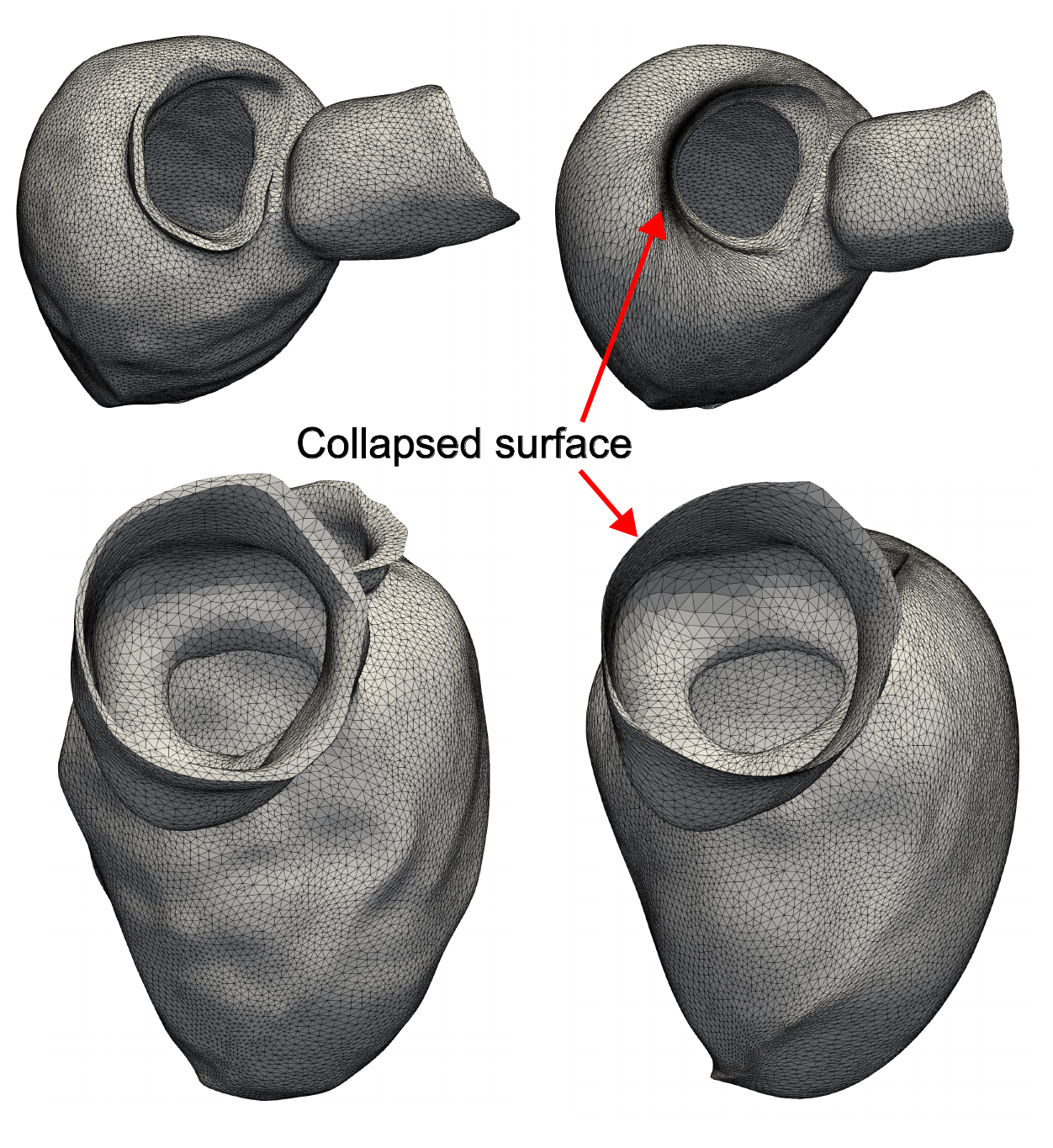}
    \caption{LT-FL (right) collapses the inlet/outlet surface. LT-FL-V (left) preserves the added thickness.}
    \label{fig:lv-collapse}
\end{figure}

\begin{figure}[t]
    \centering
    \includegraphics[width=0.47\textwidth]{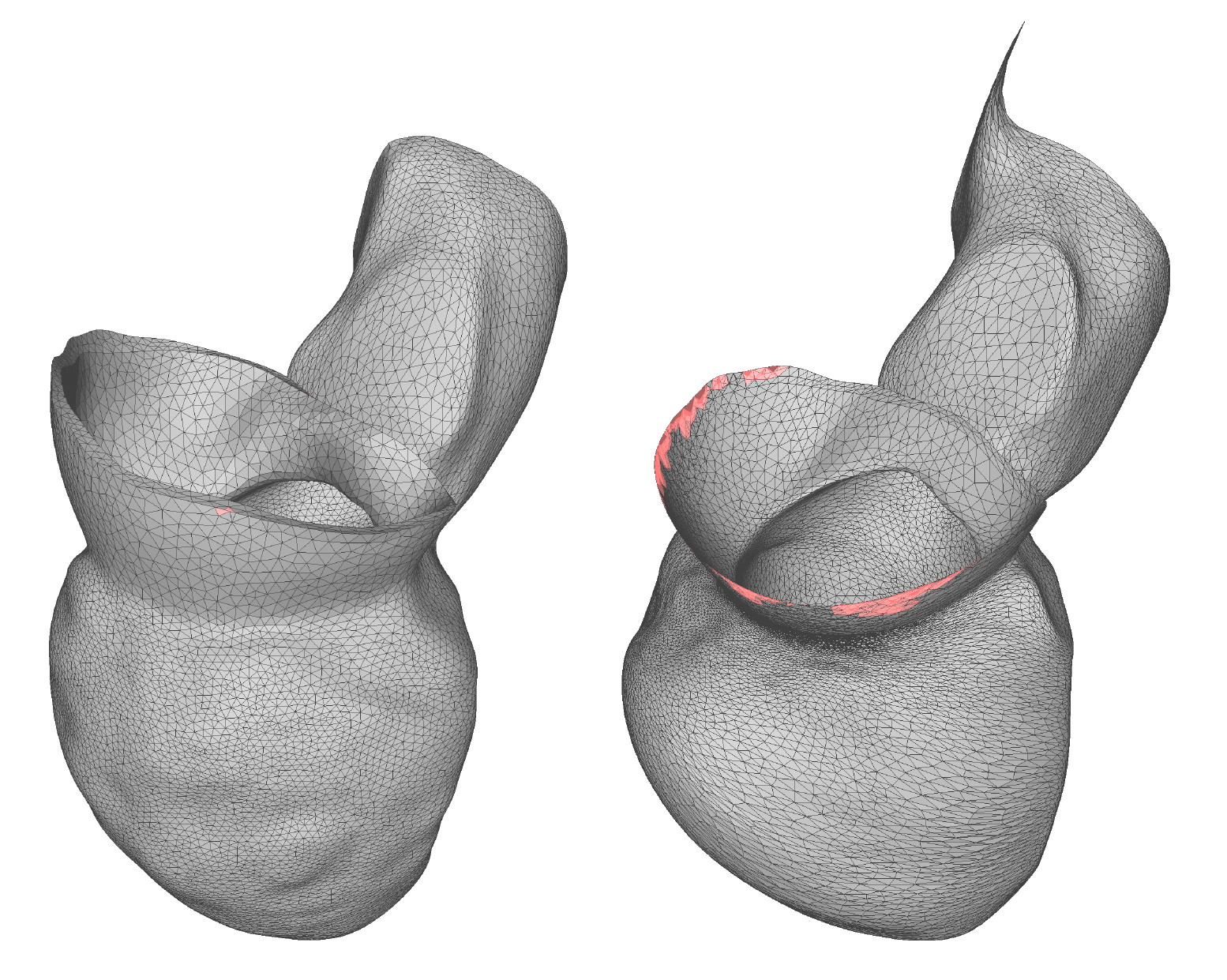}
    \caption{LT-FL-V (left) produces SIF primarily due to element collapse. LT-FL produces SIF due to surface interpenetration and element collapse. SIFs are highlighted in both images.}
    \label{fig:lv-sif-compare}
\end{figure}

An important distinction between LT-FL and LT-FL-v is the kinds of SIFs they produce. Recall that SIF can result due to surface interpenetration or element inversion. Element inversion can often be fixed quite easily using standard surface remeshing techniques. Surface interpenetration is a significantly harder problem to fix and requires manual intervention or sophisticated contact detection algorithms. While the volume loss is effective at mitigating both of these types of SIF, it is particularly effective at mitigating surface interpenetration. Figure \Ref{fig:lv-collapse} illustrates a few test samples in which the LT-FL model collapses the thickness added to the surfaces of inlets and outlets of the LV resulting in significant surface interpenetration. The LT-FL-v model is able to preserve this thickness. Figure \Ref{fig:lv-sif-compare} highlights the SIF in one of our test samples. We clearly see the different modalities of SIFs present in the meshes generated by the two models.

\begin{figure}[t]
    \centering
    \includegraphics[width=0.47\textwidth]{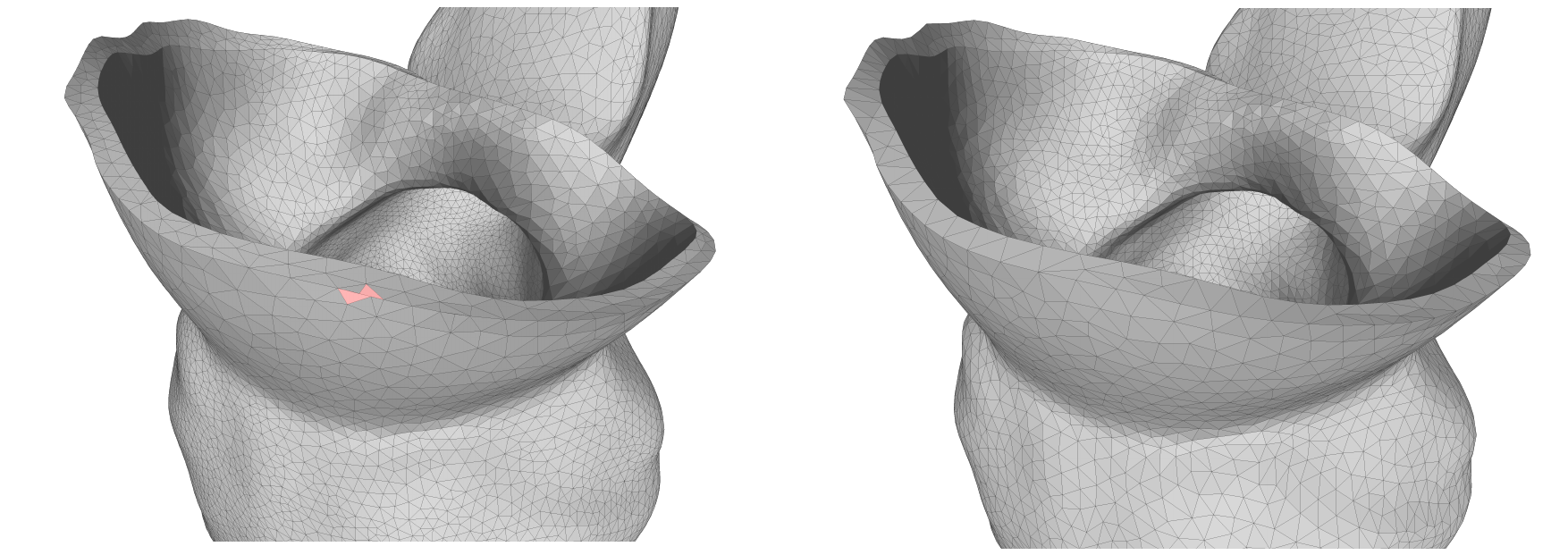}
    \caption{SIF due to element collapse (left) can be easily fixed by standard surface remeshing algorithms. Here we demonstrate the result of using the isotropic remeshing available in MeshLab.}
    \label{fig:lv-sif-fixed}
\end{figure}

Since the SIF produced by the LT-FL-v model are due to element inversions, we are able to easily fix these issues in \emph{all} of the cases using a simple isotropic remeshing available in MeshLab \cite{cignoni2008meshlab}. We illustrate how remeshing can be used to fix element collapses in fig. \ref{fig:lv-sif-fixed}. This strategy does not work for the meshes generated by the LT-FL model since the SIFs produced by this model are fundamentally different and not fixable by just remeshing. Thus, considering remeshing as a simple post-processing step, we are able to robustly generate simulation ready patient-specific meshes of the LV in \emph{all} of our test samples.


\section{Discussion} \label{sec:discussion}

We observe that HeartDeformNet \cite{kong2021deep} produces more accurate meshes when measured by the similarity metrics in Tables \ref{tab:ct-accuracy-comparison} and \ref{tab:mr-accuracy-comparison}. However, as Tables \ref{tab:ct-sif-comparison} and \ref{tab:mr-sif-comparison} show, these meshes contain non-zero self-intersections that need to be handled by post-processing. In contrast, our proposed method demonstrates comparable accuracy while robustly producing meshes with \emph{zero} self-intersections. The generated surface meshes can be readily used to produce full 3D volumetric meshes. Note that HeartDeformNet uses up to 3 mesh deformation blocks using graph convolutional networks. Similar to the results in \cite{lebrat2021corticalflow}, we anticipate that using a second flow deformation module in our workflow can help to improve our accuracy.

We demonstrated that our model trained on Whole-Heart mesh generation, may be deployed on a completely new template mesh of the myocardium along with its inlets and outlets that was not seen during training. By training our model with the volume loss eq. \ref{eq:volume-loss}, we are able to generate meshes that maintain the thickness of cardiovascular tissue for which no ground-truth data is available. Our model robustly generates meshes with a very small number of SIF. Further, these SIFs are primarily due to element inversion/collapse as opposed to surface interpenetration, and may be easily fixed using standard surface remeshing algorithms. The volume loss is effective at regularizing unphysical deformations of the template mesh.

For cardiac structures where tissue thickness is not discernible, the model cannot be trained to match tissue thickness. However, the model is trained to match the lumen surface (inner boundary of the tissue) to the image data. For such structures, a tissue thickness can be prescribed in the template, which will deform according to the flow field that deforms the lumen to the image. While our volume loss constraint prevents collapse of the tissue thickness, hence preventing surface interpenetration, it does not directly guarantee a particular tissue thickness value in the final deformed configuration. Future work may explore further constraints during the deformation process so that tissue thickness in the final configuration matches a prescribed distribution.

The kinematics of continua is a mature discipline with a rigorous and rich theoretical base. We have seen that deep learning models can be constrained by our insights from continuum mechanics. The volume loss eq. \ref{eq:volume-loss} acts as a powerful physics-based regularization for the neural network. It is a robust metric to determine if a deformation field is physically realistic. By optimizing for this measure, our model is encouraged to learn deformations that are physically realistic resulting in meshes that are more immediately usable. Numerous relations similar to eq. \ref{eq:volume-loss} can be found in the continuum mechanics literature, e.g. rate of change of lengths, areas, and normal vectors. For instance, the rate of change of an area element $A(t)$ with normal vector $n$ due to deformation by a flow vector field $v$ is,

\begin{align}
    \frac{\mathrm{d}A}{\mathrm{d}t} = A(\mathrm{div}(v) - n^T \nabla v \ n) \label{eq:area-change}
\end{align}

\noindent We believe that a loss function based on \ref{eq:area-change} can be used to prevent SIF due to element collapse as it effectively penalizes collapse modes in the plane of an element. This approach would be useful in situations where repeated remeshing is undesirable, e.g. time dependent motion of a mesh.

Template based methods have shown great promise in whole-heart mesh generation. However, there are some limitations. Firstly, templates enforce a specific topology onto the mesh. A given template is thus restricted in applicability to a given class of cardiac morphologies. When the morphology differs significantly e.g. due to congenital heart defect, we can no longer use the same template mesh. There are different solution approaches in these cases. We could consider the use of a library of template meshes depending on the target morphology. Alternatively, we could consider geometric representations, such as signed distance functions, that are able to handle changes in topology. Further, template based methods may not be effective in generating vascular meshes wherein there can be significant differences in geometry and topology due to branching and it is likely that a different strategy is required for vascular mesh generation. Our future work is focused on combining cardiac and vascular mesh generation to unify these two components into a single model.

\section{Software} \label{sec:software}

LinFlo-Net is currently being maintained at the following repository: github.com/ArjunNarayanan/LinFlo-Net. We are documenting the repository and adding usage examples after which the repository will be made available to the public.

\begin{acknowledgment}
We gratefully acknowledge the National Science Foundation (Award 1663671) for partial support of this work.
\end{acknowledgment}

%
% ...

\bibliographystyle{asmems4}

\bibliography{asme2e}



\end{document}